\documentclass[letterpaper, 10 pt, conference]{ieeeconf}  

\IEEEoverridecommandlockouts                              

\usepackage{graphicx} 
\usepackage[percent]{overpic}
\usepackage{amsmath} 
\usepackage{color}
\usepackage{soul}
\usepackage{textcomp}
\usepackage[textstyle]{SIunits}
\usepackage{breqn}

\title{\LARGE \bf
An Experimental Evaluation of Robustness and Precision for Long-term LiDAR-based Localization in Highly Changing Environments
}

\author{Salvador Dominguez$^{1}$, Ga\"{e}tan Garcia$^{1}$, Vincent Fr\'{e}mont$^{1}$, Arnaud Hamon$^{1}$ 
\thanks{$^{1}$ These authors are with LS2N, Laboratoire des Sciences du Num\'{e}rique de Nantes, Centrale Nantes and CNRS, 1 rue de la No\"{e}, 44321 Nantes, France}%
}

\begin{document}

\maketitle
\thispagestyle{empty}
\pagestyle{empty}

\begin{abstract}

  One of the hardest challenges to face in the development of a non GPS-based localization system for autonomous vehicles is the changes of the environment. LiDAR-based systems typically try to match the last measurements obtained with a previously recorded map of the area. If the existing map is not updated along time, there is a good chance that the measures will not match the environment well enough, causing the vehicle to lose track of its location. In this paper, we present and analyze experimental results regarding the robustness and precision of a map-matching based localization system over a certain period of time in the following three cases: (1) without any  update of the initial map, (2) updating the map as the vehicle moves and (3) with map updates that take into account surrounding structures labeled as "fixed" which are treated differently. The environment of the tests is a busy parking area, which ensures drastic changes from one day to the next. The precision is obtained by comparing the positions computed using the map with the ones provided by a Real-Time Kinematic GPS system. The experimental results reveal a positioning error of about $\boldsymbol{\sigma}$=\unit{6}{\centi\meter} which remains stable even after 23 days when using fixed structures on the working area.   

\end{abstract}


\section{INTRODUCTION}\label{Introd}

For autonomous navigation, it is of great importance to self-localize with good precision and high reliability in a changing environment in the long term, and this is the main motivation of this study. 

As explained in section \ref{hardware}, in the experiments we use the measurements of a 3D LiDAR that are projected onto the horizontal plane in order to simplify the processing while keeping enough information. The 2D projected scan is called \textit{occupancy vector} (see section \ref{occupancy_vector}) and is used to build occupancy grid maps.

The chosen environment for the experiments is a busy parking area on Centrale Nantes campus, in which the configuration of parked cars changes almost completely over a day. In order to test the robustness of our system, we intentionally use the LiDAR measurements in the low-height range, so that the parked cars are clearly seen on the scan. The range of height used is between \unit{0.5}{\meter} and \unit{1.7}{\meter}, so they include many parked cars, while rejecting most of the ground points. This strategy obviously leads to a high percentage of measurements not matching the map. Again, this is something one would normally try to avoid, but we are looking for an extreme and challenging case for our localization system.

Prior to the localization phase, the maps are built using Simultaneous Localization and Mapping (SLAM). Highly effective LiDAR-based SLAM techniques exist, and state-of-the-art SLAM solvers are now available that achieve good accuracy in real-time (\emph{e.g.} \textit{GMapping} \cite{Grisetti2007}, \textit{Hector SLAM} \cite{KohlbrecherMeyer2011}). We have implemented a SLAM technique based on scan matching and geo-referencing using  RTK-GPS (Real Time Kinematic GPS) which is explained in section \ref{map_building}. It is important to point out that the RTK-GPS is not a required component of the localization system. In the present context, it allows us to build precisely geo-referenced maps. Thus, the result of subsequent map-based localization can be compared to RTK-GPS positioning to evaluate precision.

For localization, our algorithms use as inputs odometry and 3D point clouds. We have built a Large Scale 2D SLAM, which uses multiple geo-referenced tile-maps organized in a grid. The tiles around the vehicle's position that cover the whole scan are loaded into memory and used to build the current sub-map for localizing the vehicle. The 3D scan is projected onto the 2D horizontal plane, generating the occupancy vector, and then matched against the sub-map to obtain the most likely position. 

When updating the map, the occupancy value of the cells of the sub-map hit by a laser measurement is increased using a \emph{delta updating value}, while for the ones that are not hit but lie between a laser measurement and the vehicle's position it is decreased by the same delta value. If a fixed structure mask is used during the map update process, a cell marked as "fixed" is not updated, even though it is hit by a laser measurement. With this technique we ensure that fixed structures remain at the same position in the map, thus avoiding long term drift. 

The objective of this work is to evaluate how updating a map along time affects the precision and stability of the localization. We also want to check whether explicitly marking some features of the map as fixed helps in the long-term updates of the map. As shown in the experimental results section, this solution ensures stable localization performance and a stable map. 

The remainder of this paper is organized as follows. In Section \ref{RelWork} we make a quick review of some related literature. Section \ref{hardware} presents the hardware configuration used for the experiments. Section \ref{map_building} contains the methodology for building and geo-positioning the tile-maps, while section \ref{map_update} explains the map update methods. Sections \ref{setup} and \ref{results} respectively present the experiments and the statistical localization results.

\section{RELATED WORK}\label{RelWork}

All environments evolve over time, and some change very rapidly. Car parking areas are examples of environments where large changes occur in a matter of hours. Others, \emph{e.g.} areas with a lot of vegetation, change in a matter of months. For dynamics of just seconds or less, there are techniques like DATMO (Detection And Tracking of Moving Objects) \cite{AzimAycard2012} that detect and label the objects around as moving or stationary, so the objects that move can be filtered out from the localization process. Similarly, SLAMMOT combines SLAM and DATMO so the map used for localization only contains the stationary information \cite{WangThorpeThrun2007}. SLAM systems are based on building a map of the environment, while simultaneously localizing the vehicle within the map. In practice, we want to be able to start with an initial map, that can be built using classic SLAM techniques, and update it progressively with new information during further passes. In the literature this problem has been addressed in different ways. In \cite{ShaikLiebig2017}, it is proposed to create an initial static map which is later updated using a temporal map created using the laser measurements, whenever the robot detects changes in the environment. In \cite{FabrizioBasilio2010} the authors present a methodology to detect variations in the environment, then they generate a local map containing only the persistent variations, using a technique called \textit{recency weighted averaging} and align it with the static map using the \textit{Hough transform}, and finally both are merged together to update the map. Other authors \cite{Milstein2005} propose an augmentation of the classical Monte Carlo Localization by applying ray-tracing on the most likely particle's position using the sensor readings while updating the map cell's probability and labeling them as observed. Although the problem of map updating has been widely addressed in the literature, there are not many studies that deal with the consequences that this update may cause in the accuracy and stability of the location in the long term.

\section{HARDWARE SETUP}\label{hardware}

\subsection{Vehicle and sensors}
\begin{figure}[!t]
      \centering
      \includegraphics[width=0.8\linewidth]{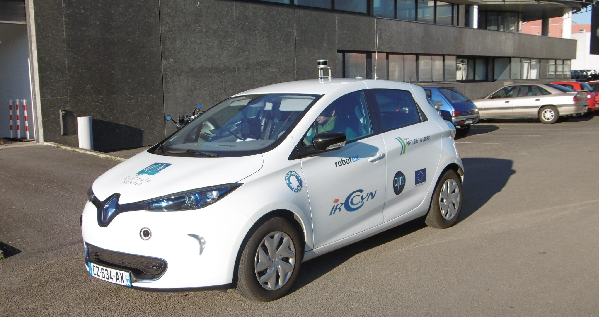}
      \caption{Vehicle used for the experiments}
      \label{Zoe}
   \end{figure}

The vehicle used for this study is an electric Renault Zoe ZE (Fig. \ref{Zoe}), equipped with the following sensors:
	  \begin{itemize}
	  \item The car's CAN bus provides individual wheel speeds at \unit{50}{\hertz}. These are used to compute the speed of the vehicle, which is one of the inputs of the odometry generator.
	  \item ``Strap-down'' Xsens MTI~100 Inertial Measurement Unit. It provides the angular speed of its three orthogonal axes at \unit{200}{\hertz}. The vertical angular speed is used to compute the increments of angular rotation of the car, which is the second input of the odometry generator.
	  \item ProFlex~800 RTK-GPS receiver. Provides positions with centimeter-level accuracy in RTK mode. It is used for ground truth generation and sub-map positioning.
	  \item Puck Velodyne VLP 16 LiDAR. Provides range measurements in 16 planes at different pitch angles between -15$\degree$ and 15$\degree$. Its horizontal coverage is 360$\degree$, with a resolution of 0.25$\degree$. The maximum range is \unit{100}{\meter}, with a range accuracy of $\pm$\unit{3}{\centi\meter}. It is used for map building and localization.
	  \end{itemize} . 

\subsection{Occupancy vector generation}\label{occupancy_vector}

The occupancy vector is the vector of points corresponding to laser measurements converted into XY coordinates, XY being an horizontal plane. The occupancy vector is expressed in the vehicle's local frame, and also stores pose covariance information. The raw data provided by the VLP-16 are 3D measurements. The 2D projection decreases the computational complexity of the matching, allowing the algorithm to run in real time on a standard computer. Nevertheless, the occupancy vector retains more information than a 2D laser scan as for the same azimuth angle there can be several associated range readings.

\begin{figure}[thpb]
      \centering
      \framebox{\parbox{8cm}{\centering \includegraphics[width=0.8\linewidth]{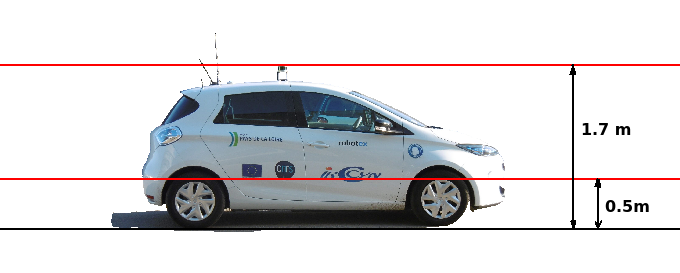}}}
      \caption{Range of heights used in the experiments for filtering the 3D point-cloud generated by the VLP-16. The points within the range of heights are selected for the generation of the occupancy vector.}      
      \label{ZoeScanLow}
\end{figure}

In the reported experiments, the occupancy vector retains the points belonging to the height range \unit{[0.5,1.7]}{\meter} (Fig.~\ref{ZoeScanLow}). Most of the points in the chosen range fall within the height of parked cars. In \cite{NobiliDominguez} a comparison between using low and high scan configurations for map localization is presented.

The VLP-16 LiDAR sensor has been placed on the roof of the car (Fig. \ref{Zoe}). Extrinsic calibration of the sensor provides the position of the VLP-16 relative to the car frame.  We convert each measurement from the sensor frame into the car frame by simple reference frame transformation. The global position of each point in the occupancy vector is computed by multiplying its position in the car's frame by the corresponding transformation of the car's global position.

\subsection{Ground truth generation}

 The ground truth position is computed using a Kalman filter that uses:
        \begin{itemize}
          \item The odometry pose. It is the result of integrating over time the horizontal speed of the car and its angular velocity around Z.
          \item The RTK-GPS position provided directly by the GPS receiver.
        \end{itemize}
        
The GPS receiver provides the position of its antenna at 1~Hz and the odometry is generated at \unit{50}{\hertz}, frequency of the speed measurements on the CAN bus of the car. The Kalman filter generates ground truth localization at the same frequency than the odometry (\unit{50}{\hertz}). 

\section{MAP BUILDING METHODOLOGY}\label{map_building}

In our localization method, the world is divided into a grid of occupancy grid tiles called \emph{map tiles}. Every map tile is further divided into a grid of square cells of a certain size. A size of 30~m~x~30~m is used for the map tiles in our experiments, with a cell size of \unit{10}{\centi\meter}~x~\unit{10}{\centi\meter}. Every cell can store a value between 0 and 100 that represents occupancy, with 0 corresponding to free space and 100 to occupied space, any other value in between means that some measures have revealed that the cell was occupied while others indicated otherwise, the result being interpretable as a (scaled) probability of the cell being occupied. Intermediate values occur, as there are many sources of noise in the process of determining the cell to which a measurement corresponds, like sensor measurement noise and frame geo-positioning error. 

\begin{figure}[thpb]
      \centering
      \includegraphics[width=0.8\linewidth]{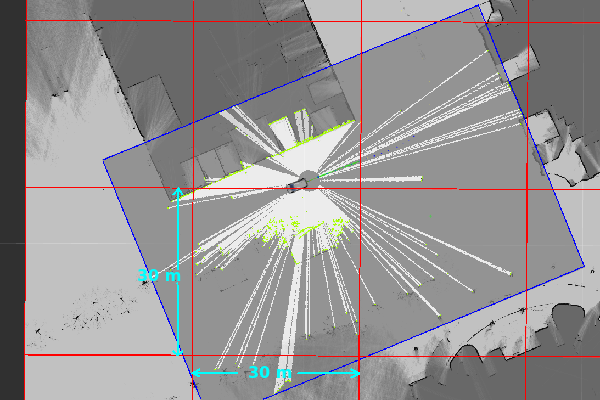}
      \caption{Map tiles in red and sub-map frame in blue during map building}
      \label{TileMapBuilding}
   \end{figure}

During map building, when a new occupancy vector is generated from a scan, it is synchronized with the pose provided by the odometry+GPS Kalman filter, by interpolating the position to the time stamp of the scan. The next step is to build an occupancy grid map frame (blue rectangle in Fig.~\ref{TileMapBuilding}) with that scan. Its orientation is that of the car and it contains all measurement points. It is then matched with the previous frame in a buffer of up to 20 frames. When the number of frames in the buffer is enough, we apply a map positioning optimization method explained in more detail in \cite{ITSC2015MultipleMaps}, using the RTK-GPS positions of the frames in the buffer and their respective positions after inter-frame matching. With the optimized positions we build an intermediate \emph{sub-map} by projecting all the frames of the buffer onto a single occupancy grid map. Since the position and orientation of the sub-map does not match that of the map tiles, it is then re-projected onto the map tiles (with red edges in Fig.~\ref{TileMapBuilding}) using the corresponding coordinate transformation.

This method aims at maintaining a good matching between consecutive frames, while being globally consistent as it uses the RTK-GPS position.

\begin{figure}[thpb]
\centering
  \includegraphics[scale=0.45]{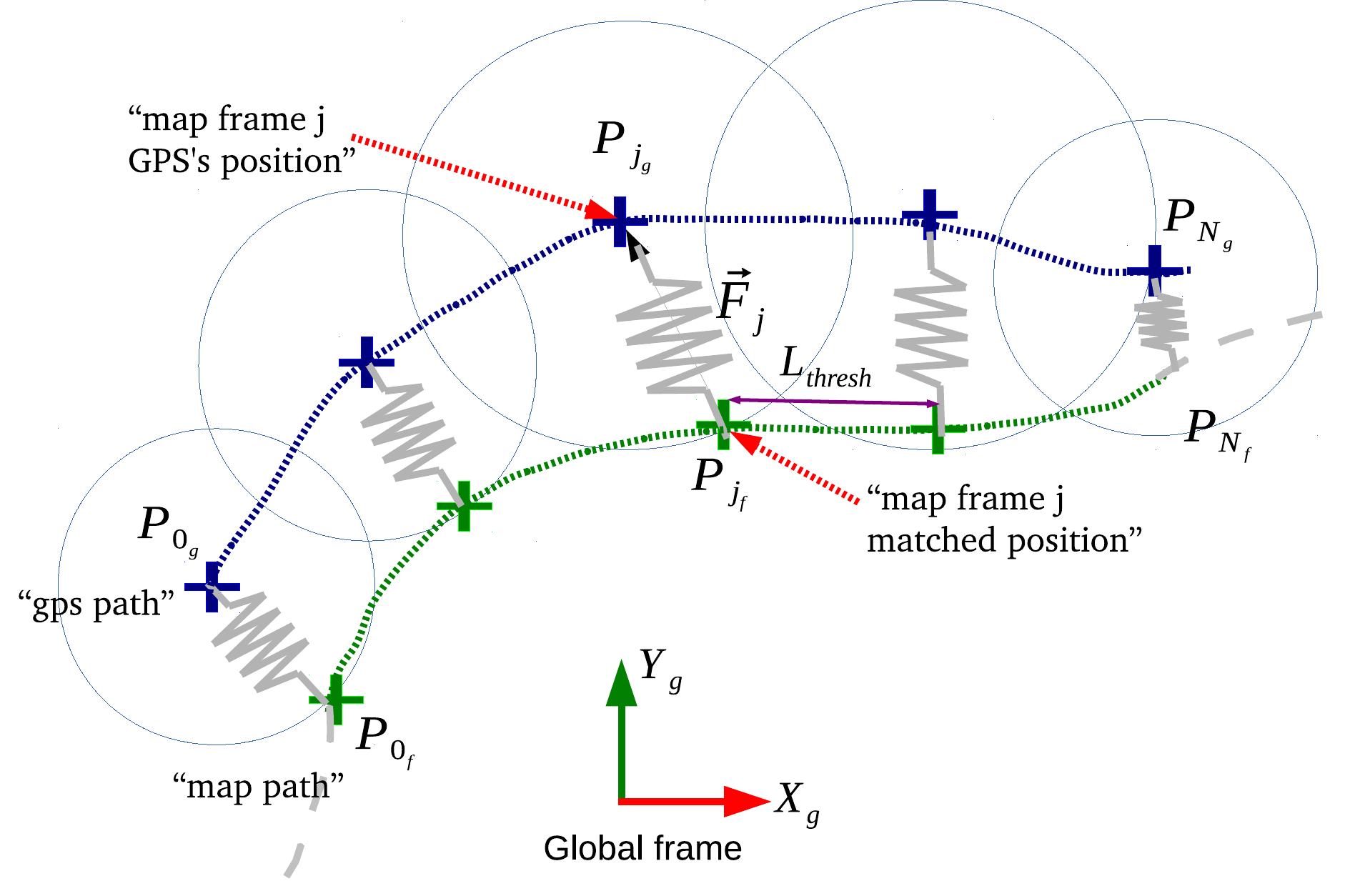}
  \caption{Every frame is attracted from its matched position $P_{j_f}$ towards the corresponding GPS point $P_{j_g}$ with a force ${\vec F}_j$ proportional to the inverse of the covariance of the position error $cov_j$ (represented by the blue circles) and the distance between them.}
  \label{MapFrameGeoPositioning}
\end{figure}

The map frames are positioned forming a chain in which two consecutive frames are matched together. Each pair, map frame position $\mathcal{P}_{f_j}$ and its corresponding Kalman filter position $\mathcal{P}_{g_j}$ are recorded together for the same time instant. We also record the variance of the localization error of the Kalman filter solution $\sigma^2_{g_j}$. In this way we obtain two paths, the \emph{map frames path} and the \emph{GPS path} (Fig.~\ref{MapFrameGeoPositioning}), produced respectively by the scan matching and Kalman filter processes. We consider that both paths are not deformable during the geo-positioning process. The objective of the optimization by relaxation process for geo-positioning is to minimize the energy of fitting together both paths. The output of the relaxation process is the sequence of map frame poses so that the coherence of the information recorded on the global path and the map frame path are optimal. 

As mentioned before, this process is only applied when the number of map frames in the buffer reaches a maximum size, so it takes little computing resources and can be run in real time on a standard computer. 

\section{MAP UPDATE METHODOLOGY}\label{map_update}

An initial map building phase is run in order to generate the map tiles covering the whole working area. These tiles are geo-referenced square maps that represent the "probability" of occupancy in the area covered by the map with a given resolution. At every scan, all the tiles that cover the scanned area around the vehicle's position are loaded into memory and kept there during a minimum number of scans. A rectangular earth-axes oriented sub-map is built using the tiles in memory. This is the sub-map used for localizing the vehicle at any moment. This sub-map changes in position and size when the car moves, as new tiles are loaded into memory and others are removed from it. The scan occupancy vector is then placed at a random number of poses around the last estimated pose, following a Gaussian distribution. For each pose, a matching index is computed (Eq. \ref{eq1}) and the pose which has the highest matching index is retained. The process is performed after a given distance has been traveled or a certain time has elapsed, whichever comes first. The obtained pose is used as a measurement in a Kalman filter whose prediction step is odometry at \unit{50}{\hertz}, just as we use GPS measurements when calculating the ground truth.    

The expression used to compute the matching index of a scan occupancy vector is:
\begin{dmath}\label{eq1}
match=0.5+\frac{1}{2\left(N_a+\alpha N_b\right)}\left(\sum^{N_a-1}_{i=0}{{\left|{{\frac{occ_i}{50}}-1}\right|}.{\left({{\frac{occ_i}{50}}-1}\right)}}+
\alpha\sum^{N_b-1}_{j=0}{{\left| {{\frac{occ_j}{50}}-1}\right|}.{\left( {{\frac{occ_j}{50}}-1}\right)}}\right)
\end{dmath}

where $N_a$ is the number of measurements that do not fall into fix structure cells, $N_b$ the number of measure that do, $\alpha$ is the fixed structure cell weight, $occ_i$ or $occ_j$ is the occupancy value of the cell $i$ or $j$. Note that the term in the summations can only adopt values in the range [-1, 1]. The sum of $N_a$ and $N_b$ is the total number of measurements in the occupancy vector. $\alpha$ is a value greater than 1 in order to give more importance to fixed structure cells, so that the best scan matching pose tends to fit more intensely on cells of fixed structures than on the rest.

\subsection{Map updating without fixed structures}\label{map_update_normal}

Eq.~\ref{eq1} represents the general case where fixed structures are used. In the present case we get rid of those structures by setting $\alpha=0$ in that equation and $N_a$ includes all the measurements in the occupancy vector.
For map updating we first build the occupancy grid map frame that encloses the current occupancy vector. The cells that correspond to the laser measurements are marked as occupied (value = 100), while the cells in between the car's position and those measurements are marked as free (value = 0) and the rest as unknown, like for the map-frame in Fig.~\ref{TileMapBuilding} (in blue). For every cell in the map frame we check its value and modify the value in the underlying map tile following equation (\ref{eq2}):

\begin{dmath}\label{eq2}
  v_{tile}=\begin{cases}
    \max(v_{tile}-\delta_{updt},0), \text{if $\left(v_{frame}=0\right)$} \\
    \min(v_{tile}+\delta_{updt},100), \text{if $\left(v_{frame}=100\right)$}
   \end{cases}
\end{dmath}

where $v_{tile}$ and $v_{frame}$ are the values of the cell in the tile map and map frame respectively and $\delta_{updt}$ is the increment added or subtracted from the depending on whether the cell is free or occupied in the map frame. Experimentally a good value of $\delta_{updt}$ is one that neither, increases very slow nor very fast which depends also on the update rate. In our case we update every \unit{5}{\meter} with $\delta_{updt}=6$.

\subsection{Map updating with fixed structures}\label{map_update_fix_structures}

\begin{figure}[thpb]
\centering
  \includegraphics[width=0.8\linewidth]{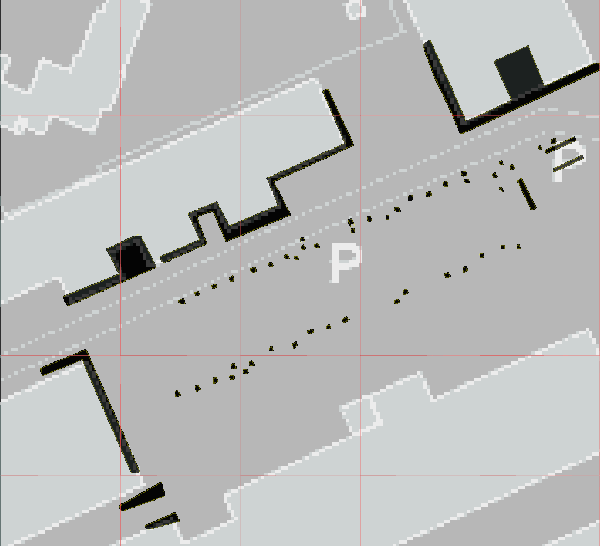}
  \caption{Map mask of fixed structures (represented as black areas).}
  \label{UpdateMaskMap}
\end{figure}

In order to distinguish which cells belong to fixed structures and which don't, we define a map mask (currently a manual process) as shown in Fig. \ref{UpdateMaskMap}. The mask is a binary map, with one value representing areas where structures are know to be fixed.

The rule for updating the cell of a map tile is still represented by Eq.~\ref{eq2}, except that only cells which are not marked as fixed are updated.

\section{EXPERIMENTAL CONDITIONS}\label{setup}

\begin{figure}[thpb]
\centering
  \includegraphics[width=0.8\linewidth]{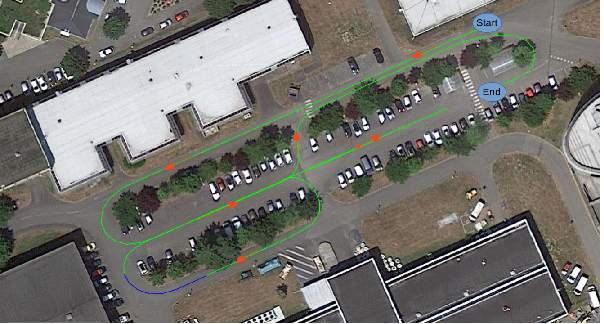}
  \caption{Experimental site and path followed for the tests.}
  \label{ParkingECN}
\end{figure}

For the experimental tests we have chosen a busy parking area on Centrale Nantes campus (Fig.~\ref{ParkingECN}) in which the datasets were recorded at different dates and times, so we can ensure a different distribution of cars in all cases. 
The total length of the test is about \unit{540}{\meter} and it passes twice along the central alley, between the two rows of cars parked on both sides. This is the most critical part of the test route, as in many cases the sensor will detect mostly parked vehicles and very few fixed features of the environment.

Three cases have been considered for comparison:
\begin{itemize}
 \item Case~1: Localization using the map of the first day (no map update).
 \item Case~2: Localization starting with the map of the previous day. The map is updated without explicitly marking any structure as fixed.
 \item Case~3: Localization using the map of the previous day while updating the map, except for areas marked as fixed in the map mask. 
\end{itemize}

Remarks about the cases: 
\begin{itemize}
 \item Case~1: The map is built using the dataset of the first day (the parking area was mostly empty). All tests use that map without any modification.
 \item Case~2: The test starts with the map updated during last test, and the map is updated during the test. The resulting map is used at the start of the next test. 
 \item Case~3:
 	\begin{itemize}
 		\item As for case 2, the map resulting of the previous test is used for localization, while updating it. The resulting map is used the following day.
 		\item We have labeled as fixed the following items: tree trunks close to parked cars, some buildings around the parking area, street lights and posts. 
	\end{itemize}
\end{itemize}

\section{RESULTS AND ANALYSIS}\label{results}

\begin{figure}
\centering
\begin{overpic}[scale=0.37]{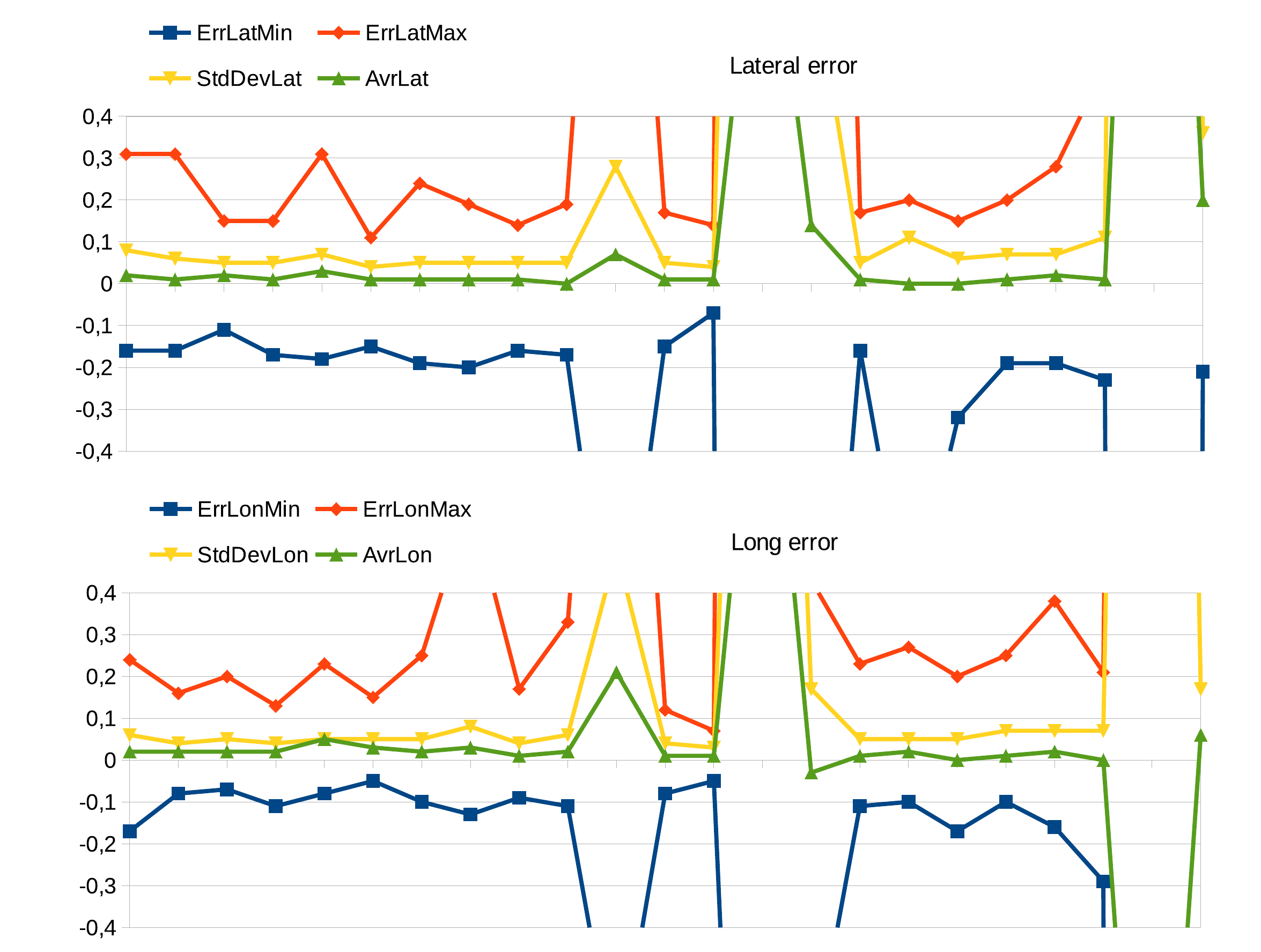}\end{overpic}
\caption{Case 1: Lateral and longitudinal error statistics of all datasets. Horizontal axis: datasets in chronological order. Error values are in meters.}
\label{fig:FirstScenarioPlots}
\end{figure}

\begin{figure}
\centering
\begin{overpic}[scale=0.42]{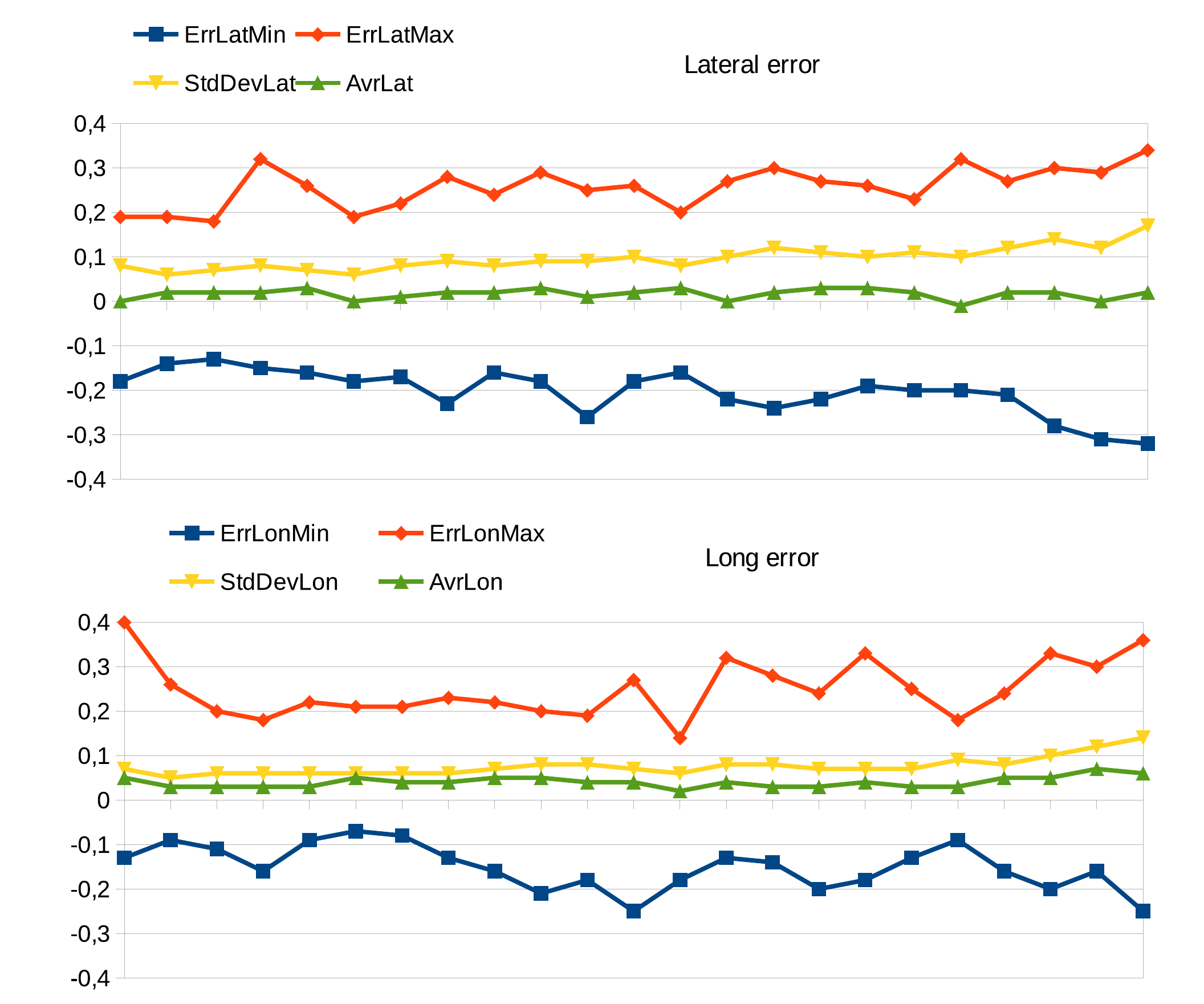}\end{overpic}
\caption{Case 2: Lateral and longitudinal error statistics of all datasets. Horizontal axis: datasets in chronological order. Error values are in meters.}
\label{fig:SecondScenarioPlots}
\end{figure}

\begin{figure}
\centering
\begin{overpic}[scale=0.42]{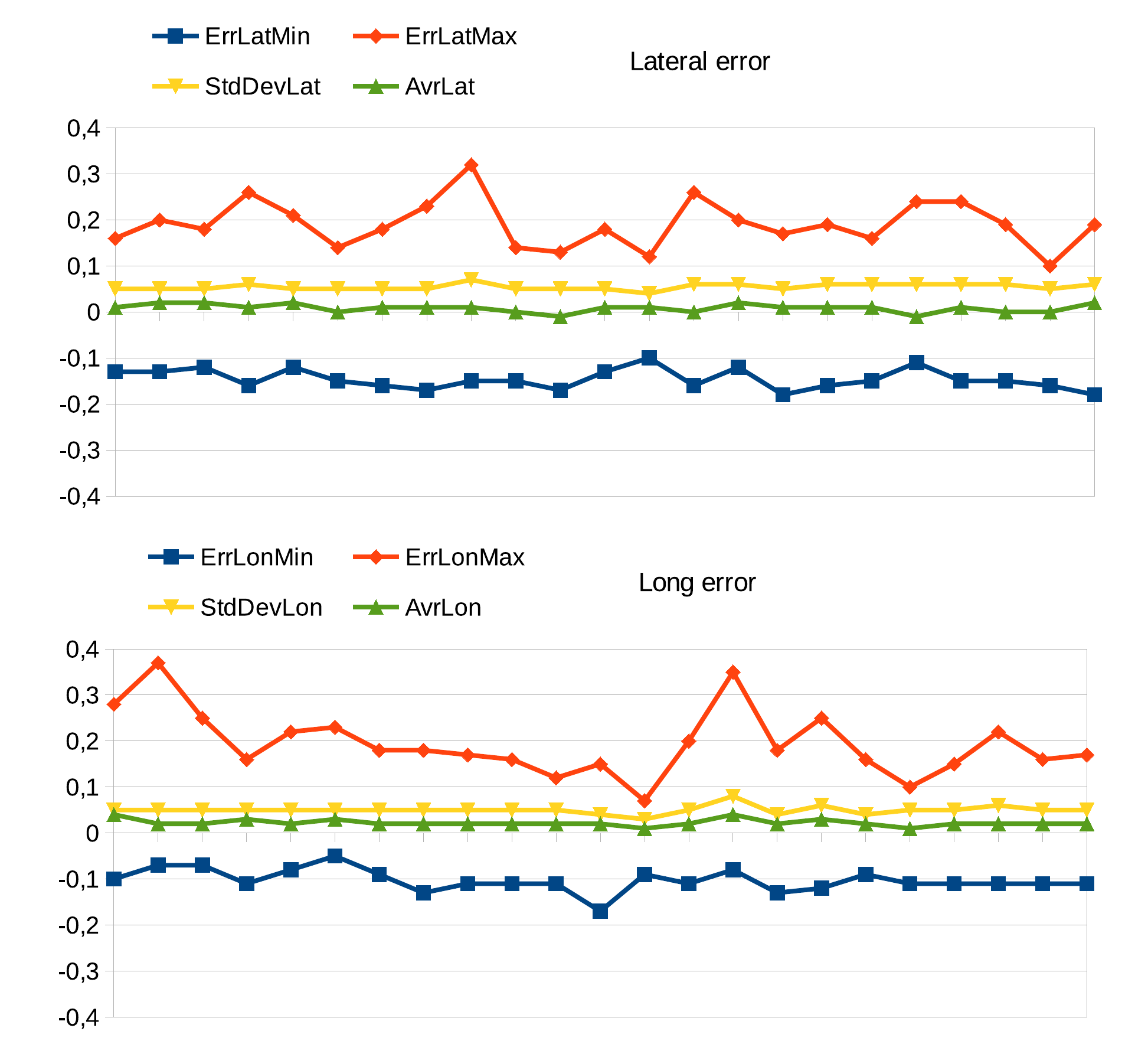}
\end{overpic}
\caption{Case 3: Lateral and longitudinal error statistics of all datasets. Horizontal axis: datasets in chronological order. Error values are in meters.}
\label{fig:ThirdScenarioPlots}
\end{figure}

Fig.~\ref{fig:FirstScenarioPlots},~\ref{fig:SecondScenarioPlots}, and~\ref{fig:ThirdScenarioPlots} show the error statistics for the three cases, run over the same data collected on 23 days. Each test being \unit{540}{\meter} long, the total traveled distance of \unit{12.42}{\kilo\meter}. The graphs show the minimum and maximum errors, the average error and its standard deviation for each dataset.


\begin{figure}
\centering
\begin{overpic}[scale=0.4]{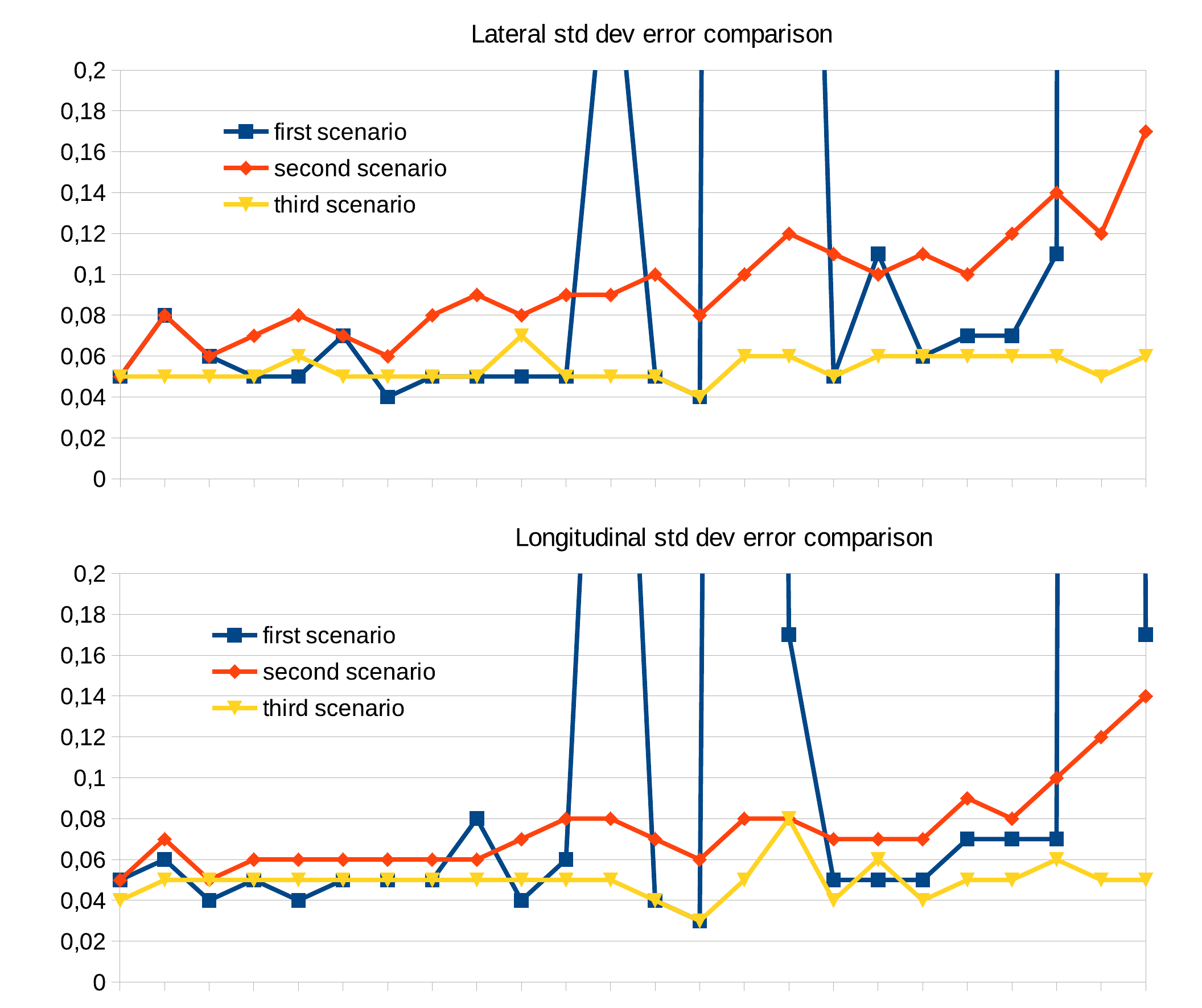}
\end{overpic}
\caption{Standard deviation for the three cases under analysis. Note the progressive increase of the std. dev. of the second case. Horizontal axis represents the datasets in chronological order, and error values are in meters}
\label{fig:StdDevComparisonPlots}
\end{figure}

Fig. \ref{fig:StdDevComparisonPlots} plots the lateral and longitudinal standard deviations of the errors of the three cases. It shows the fundamental behavior differences between them. With case~3, the error standard deviation is stable. On the contrary, case~2 shows a slow but steady increase of the standard deviation (red plot). Case~2 is never lost in these tests, but successive updates of the map cause a long term instability. 

The situations where case~1 obtains similar results to case~3 correspond to days/times when the parking is not full, as was the case with the initial map. But case~1 can fail badly, with errors off the chart, in a full parking situation. Closer analysis shows that it always happens in the same areas, where the sensor hardly detects any of the unmarked fixed structures of the environment.

Here are a few observations we consider important:
\begin{itemize}
 \item About the max-min values, it must be considered that the error results include the GPS positioning error, which is about \unit{1-3}{\centi\meter} of standard deviation. There are also some transient effects like the inclination of the car due to its suspension while turning, which can produce some peaks on the max-min values of the error. 
 \item Case~1: in some cases the localization diverged. It always happened in particular locations of the central alley at hours in which the parking was almost full. The matching index did not exceed a predefined threshold, in which case no measurements were fed to the Kalman filter. In other cases, a mismatch happened and incorrect measurements were fed to the filter. However, on days when the parking was not too occupied the precision was comparable to case~3.
 \item Case~2: as we update the map while the car moves, we do not observe large error peaks or divergence. That reveals a good robustness against changes in the environment in the medium term. However, we observe a progressive increase in the standard deviation from about \unit{5}{\centi\meter} to about \unit{12}{\centi\meter} lateral, and from about \unit{5}{\centi\meter} to \unit{10}{\centi\meter} longitudinal. The rate of increase is thus about \unit{0.33}{\centi\meter} and \unit{0.24}{\centi\meter} per update respectively. We observe also that the max-min values have the same tendency. A consequence on the occupancy grid maps is that they become more and more blurred as the number of updates on the same area increases. A plausible explanation of this effect is that the sensor measurements have some noise which, combined with the noise of the localization itself, produces that a given point of an obstacle hit by the laser scan doesn't fall always on the same cell of the map, which progressively blurs the outline of the obstacle in the map after updating it. 
 \item Case~3: we observe a good localization stability. The errors remain under reasonable limits and the overall precision is as good as in the best results of the first case. Unlike the second case we observe a constant value of the standard deviation around \unit{5-6}{\centi\meter}, as well as for the tendency on the max-min values. The fact that the parts of the map corresponding to fixed structures are not modified during map updates prevents the localization system from drifting in terms of global position and error standard deviation along consecutive updates. We conclude that the robustness and precision are good enough for long-term navigation with this system in a changing environment. As the fixed structures are not updated in the map, they contribute to avoid drifting in the global position and consequently prevents the rest of the map from becoming burred after multiple updates as happened in case~2.

\end{itemize}

\addtolength{\textheight}{-12cm}   




\section{CONCLUSIONS AND FUTURE WORK}\label{conclusions}

Results of an intensive campaign of evaluation of a multi-map scan matching-based localization algorithm in a highly changing environment have been reported. Three different cases have been analyzed in order to compare the degree of robustness and precision obtained, where (1) the map is not updated at all, (2) it is updated at every pass and (3) it is updated at every pass but some parts of the map, labeled as fixed structures, are not. The results reveal that the definition of such fixed structures, not only maintains a constant level of precision along time, but also avoids the degeneration of the map when adding new information, thus contributing to the robustness and stability of the algorithm.

The process of definition of fixed structure masks is performed by hand at this stage of development. In the near future, we consider developing an automatic construction of the fixed structures map mask by measuring the degree of "stationarity" of each cell, so the system can decide whether to include it in the map mask or not.

\section*{ACKNOWLEDGEMENTS}

The equipment used to carry out the tests has been sponsored by the French government research programme "Investissements d'Avenir" through the Robotex Equipment of Excellence (ANR-10-EQPX-44). We would like to thank also the HIANIC project consortium for its collaboration in this paper.


\bibliographystyle{IEEEtran}
\bibliography{IEEEabrv,biblio}

\end{document}